\documentclass[journal]{IEEEtran}

\ifCLASSINFOpdf
 
\else
  
\fi

\usepackage{algorithm}
\usepackage{algpseudocode}
\usepackage{times,amssymb,amsmath,amsfonts,float,nicefrac,float,accents,color,stmaryrd,graphicx,bbm,caption,subcaption,mathrsfs,amsthm}
\usepackage{balance}

\newtheorem{theorem}{Theorem}
\newtheorem*{theorem*}{Theorem}

\newtheorem{corollary}[theorem]{Corollary}
\newtheorem*{corollary*}{Corollary}
\theoremstyle{remark}

\theoremstyle{remark}

\newcommand\remove[1]{}
\newcommand{\ba}{\begin{array}}
\newcommand{\ea}{\end{array}}
\newcommand{\be}{\begin{equation}}
\newcommand{\ee}{\end{equation}}
\newcommand{\bea}{\begin{eqnarray}}
\newcommand{\eea}{\end{eqnarray}}

\newcommand\nc\newcommand

\begin{document}

\title{Comments on ``Deep Neural Networks with Random Gaussian Weights: A Universal Classification Strategy?''}

\author
{Talha Cihad~Gulcu,~\IEEEmembership{Member,~IEEE,}
Alper~G{\"u}ng{\"o}r,~\IEEEmembership{Member,~IEEE}
\thanks{The authors are with Aselsan Research Center,
Yenimahalle, Ankara, 06370, Turkey
(e-mail: tcgulcu@gmail.com, alpergungor@aselsan.com.tr).}
\thanks{A.~G{\"u}ng{\"o}r is also with Department of Electrical
and Electronics Engineering, Bilkent University, Bilkent, Ankara,
06800, Turkey.}
%\thanks{Manuscript received 2019; revised 2019.} 
}

% The paper headers
%\markboth{IEEE Transactions on Signal Processing, ~Vol., ~No., ~2018}{T. C. Gulcu and A. G{\"u}ng{\"o}r: Comments on Deep NNs with Random Gaussian Weights}

\maketitle

\begin{abstract}
In a recently published paper \cite{main}, it is shown that deep neural networks (DNNs) with random Gaussian weights preserve the metric structure of the data,
%a distance-preserving embedding for the data points, 
with the property that the distance shrinks more when
the angle between the two data points is smaller.
We agree that the random projection setup considered in
\cite{main} preserves distances with a high probability.
But as far as we are concerned,  the relation between the angle of the data points and the output distances is quite the opposite, i.e., smaller angles result in
a weaker distance shrinkage. 
This leads us to conclude that Theorem~3 and Figure~5 
in \cite{main} are not accurate. 
Hence the usage of random Gaussian 
weights in DNNs cannot provide an ability of universal 
classification or treating in-class and out-of-class data separately.
Consequently, the behavior of networks consisting of random Gaussian weights only is not useful to explain how DNNs achieve state-of-art results in a large variety of problems.
\end{abstract}
%\vspace{-0.1in}
\begin{IEEEkeywords}
Artificial neural networks, computation theory, deep learning, learning systems.
\end{IEEEkeywords}

\IEEEpeerreviewmaketitle

\section{Introduction}
Deep neural networks (DNNs) have gained popularity in recent years thanks to their achievements in many applications including computer vision, 
%image recognition, object detection, image %segmentation,
signal and image processing,
speech recognition \cite{Krizhev12}--\cite{Szegedy17}. %,Hinton12,Farabet13,%Dong16,Szegedy17}
For the purpose of providing insights into remarkable 
empirical performance of DNNs, the paper \cite{main} focuses on the properties of deep networks with random weights.
In particular, it is proved in \cite{main} that 
the presence of random i.i.d. Gaussian weights provide a distance-preserving embedding for the data points.
In the same work, it is also stated (formally in Theorem~3) that DNN layers having random coefficients distort the Euclidean distances
``proportionally to the angles between its input points:
the smaller the angle at the input, the stronger the shrinkage of the distances.''

Given the fact that the angles between data points belonging to different classes are generally larger than those of the points within the same class \cite{Yamaguchi98}--\cite{Qiu15}, the angle versus distance shrinkage relation asserted in
\cite{main} implies ``deep neural networks with random weights are universal systems that separates any data (belonging to a low dimensional model) according to the angles between its points'', as stated in \cite{main}.
In fact this implication seems to be the main contribution of the paper \cite{main}, as evident from the words ``a universal classification strategy'' being stressed in its title.

The classification ability of DNNs with random weights is emphasized in numerous places throughout the 
paper \cite{main}. For example, the authors claim in the introduction that
``Our theory shows that the addition of ReLU makes the system sensitive to the angles between points. We prove that networks tend to decrease the Euclidean
distances between points with a small angle between
them (`same class'), more than the distances between points with large angles between them 
(`different classes')'' and ``DNN are suitable for
models with clearly distinguishable angles between the
classes if random weights are used.'' In some other
sections they have similar statements, such as ``It can be observed that the distance between points with a smaller angle between them shrinks more than the distance between points with a larger angle between them. Ideally, we would like this behavior, causing points belonging to the same class to stay closer to each other in the output of the network, compared to points from different classes'' and
``In general, points within the same class would have small angles within them and points from distinct classes would have larger ones. If this holds for all the points, then random Gaussian weights would be an ideal choice for the network parameters.'' 

The angle dependence of distance distortion which leads to an ideal classifier is mentioned in the conclusion section of \cite{main} as well, in the following lines: 
``While preserving the structure of the initial metric is important, it is vital to have the ability to distort some of the distances in order to deform the data in a way that the Euclidean distances represent more faithfully the similarity we would like to have between points from the same class. We proved that such an ability is inherent to the DNN architecture: the Euclidean distances of the input data are distorted throughout the networks based on the angles between the data points. Our results lead to the conclusion that DNNs are universal classifiers for data based on the angles of the principal axis between the classes in the data.''

It is also worthwhile to mention the papers citing \cite{main}.
Most of the papers cite \cite{main} in an undetailed fashion as one of the papers providing some theoretical analysis for DNNs 
\cite{Daniely16}--\cite{Papyan17}
or as a study showing us the distance-preserving aspect
of random DNN weights \cite{Papyan17,Sokolic17}. But still
some remarks on the classification property associated with
random weights, similarly to the ones made by \cite{main}, are encountered in some of the
papers citing \cite{main}, such as ``According to their observation, random filters are in
fact a good choice if training data are initially well-separated'' in \cite{Gilbert17}, 
``Deep networks with
random weights are a universal system that separates any
data (belonging to a low-dimensional model) according to the
angles between the data points, where the general assumption
is that there are large angles between different classes''
in \cite{Vidal17},  ``Giryes et al. (2015) proved that under random Gaussian weights, deep neural networks are distance-preserving mappings with a special treatment for intra- and inter-class data'' in \cite{Vo17},  
``Randomness of features has been used with great success
as a mean of reducing the computational complexity of neural
networks while achieving comparable performance as with
fully learnt networks'' in \cite{Venk18},  
``The authors demonstrate that this form of nonlinear
random projection performs a class-aware embedding where the
embedding places objects of the same class closer to one another
after the projection compared to objects of different classes''
and
``such embeddings tend to decrease the Euclidean distances between points with a small angle between them (`same class') more than the distances between points with large angles between them (`different classes')'' in \cite{Karimi17}, and
``The model structure of the RVFL is so simple, why does
the RVFL work well for most tasks? Giryes et. al. give
a possible theoretical explanation for this open problem.''
(here RVFL is the abbreviation for random vector functional
link) in \cite{Zhang17}.

\vspace{-0in}
In this work, we argue that the claims quoted above and
made by \cite{main} along with some of the papers citing it
in regards to the universal classification
or class-aware embedding
require reconsideration by disproving
Theorem 3 (more precisely Eq.~(4)) in \cite{main}. 
For that purpose, we consider the problem setup
of this theorem and calculate the expectation of
the squared norm term 
%$||\rho({\mathbf M}{\mathbf x})-\rho({\mathbf M}
%{\mathbf y})||_2^2$ 
appearing in the theorem statement. This would be
the subject of Section~\ref{thm3}, where we also show 
for a single layer of a randomly weighted DNN that
smaller angle values between the input pairs cause a larger output Euclidean distance to input Euclidean distance ratio, contrary to what is claimed by \cite{main}. We briefly discuss 
Theorem 4 and derive the revised form of Corollary 5 
in \cite{main}, in Section~\ref{cor5}.
Lastly, we discuss the consequences of our derivations in
conjunction with the simulation results presented by \cite{main} in Section~\ref{concl}. 
\vspace{-.22in}
\section{Problem Definition and The Revised Distance Distortion Result}
\vspace{-.1109in}
\label{thm3}
The effect of a single layer of DNN on the Euclidean distances is considered in \cite{main}, with $\rho$ being the rectified linear unit (ReLU) activation function, 
${\mathbf M}$ being the linear operator, and 
$\rho({\mathbf M}\cdot)$ accounting for the transformation applied at a single layer. Denoting
the Euclidean ball of radius $r$ by 
${\mathbb B}_r^n\subset {\mathbb R}^n$, the authors of
\cite{main} prove the following theorem.

\begin{theorem*}[Theorem~3 in \cite{main}]Let 
$K\subset{\mathbb B}_1^n$ be the manifold of the data
in the input layer. If 
$\sqrt{m}{\mathbf M}\in {\mathbb R}^{m\times n}$ is a random
matrix with i.i.d. normally distributed entries and
$m\geq C\delta^{-4}\omega(K)^4$ (here $\omega(K)$
is the Gaussian mean width defined as $\omega(K)\triangleq{\mathbb E}[\sup_{{\mathbf x},{\mathbf y}\in K}\langle {\mathbf g},{\mathbf x}-{\mathbf y}\rangle]$), then with a high probability
(of the form $1-\exp(-O(\delta^2))$) the inequality
given by 
\begin{align}
&\Bigg| 
\|\rho(\mathbf{Mx})-
\rho(\mathbf{My})\|_2^2\nonumber\\&\quad-
\left(\frac{1}{2}
\|{\mathbf x}-{\mathbf y}\|_2^2+
\|{\mathbf x}\|_2\|{\mathbf y}\|_2
\psi({\mathbf x},{\mathbf y})\right)\Bigg|\leq \delta
\label{eq4}
\end{align}
holds, where
\be
\psi({\mathbf x},{\mathbf y})=\frac{1}{\pi}
\left(\sin\angle({\mathbf x},{\mathbf y})-
\angle({\mathbf x},{\mathbf y}) \cos\angle({\mathbf x},{\mathbf y})
\right)
\label{psi}
\ee
with $0\leq \angle({\mathbf x},{\mathbf y}) 
\triangleq \cos^{-1}
\left(\frac{{\mathbf x}^T{\mathbf y}}{\|{\mathbf x}\|_2\|{\mathbf y}\|_2}\right)\leq\pi.$
\end{theorem*}
The proof of Theorem 3 in \cite{main} relies on some concentration inequality for Lipschitz--continuous functions of Gaussian random variables and Bernstein's inequality, together with the observation that
\be
{\mathbb E}\|\rho(\mathbf{Mx})-\rho(\mathbf{My})\|_2^2
=\frac{1}{2}\|{\mathbf x}-{\mathbf y}\|_2^2+
\|{\mathbf x}\|_2\|{\mathbf y}\|_2
\psi({\mathbf x},{\mathbf y}),
\label{expe}
\ee
see Eq.~(22) in \cite{main}.
The function $\psi(\theta)=\frac{1}{\pi}
(\sin\theta-\theta\cos\theta)$ is increasing on the interval
$[0,\pi]$, and consequently it follows from \eqref{eq4} that 
the smaller the angle between ${\mathbf x}$ and 
${\mathbf y}$, the smaller the output distance turns out to be. This behavior of
a single layer of DNN with random weights is also summarized in Fig.~5 of \cite{main}, where it is illustrated that two classes with distinguishable angles can be separated using such networks. 

In the rest of this section, we evaluate the expression
${\mathbb E}\|\rho(\mathbf{Mx})-\rho(\mathbf{My})\|_2^2$,
and show that \eqref{expe} (and thus Theorem~3 in \cite{main}) 
is not true. For that purpose, we
first remind the reader Eq.~(18) and Eq.~(19) in \cite{main}: 
\begin{align}
\|\rho(\mathbf{Mx})-\rho(\mathbf{My})\|_2^2&=
\sum_{i=1}^m 
\left(\rho({\mathbf m}_i^T{\mathbf x})-
\rho({\mathbf m}_i^T{\mathbf y})\right)^2\label{expe1}\\
{\mathbb E}\left(\rho({\mathbf m}_i^T{\mathbf x})-
\rho({\mathbf m}_i^T{\mathbf y})\right)^2&=
{\mathbb E}\left(\rho({\mathbf m}_i^T{\mathbf x})\right)^2+
{\mathbb E}\left(\rho({\mathbf m}_i^T{\mathbf y})\right)^2\nonumber\\&\quad
-2\,{\mathbb E}\left[\rho({\mathbf m}_i^T{\mathbf x})\rho({\mathbf m}_i^T{\mathbf y})\right].\label{expe2}
\end{align}

The terms ${\mathbb E}\left(\rho({\mathbf m}_i^T{\mathbf x})\right)^2$ and 
${\mathbb E}\left(\rho({\mathbf m}_i^T{\mathbf y})\right)^2$ appearing in \eqref{expe2} can be easily computed from the
symmetry of Gaussian distribution as 
\begin{align}
{\mathbb E}\left(\rho({\mathbf m}_i^T{\mathbf x})\right)^2
&=\frac{1}{2}\,
{\mathbb E}\left({\mathbf m}_i^T{\mathbf x}\right)^2\nonumber\\
&=\frac{1}{2}\sum_{k=1}^n {\mathbb E}({\mathbf m}_i(k))^2 
({\mathbf x}(k))^2\nonumber\\&
=\frac{1}{2m}\|{\mathbf x}\|_2^2\label{expe3}\\
{\mathbb E}\left(\rho({\mathbf m}_i^T{\mathbf y})\right)^2
&=\frac{1}{2m}\|{\mathbf y}\|_2^2,\label{expe4}
\end{align}
consistently with Eq.~(20) in \cite{main}, where
$k$ in brackets denote the $k^{\text{th}}$ component of the corresponding vector. 

\begin{figure}
    \centering
    \includegraphics[scale=0.14]{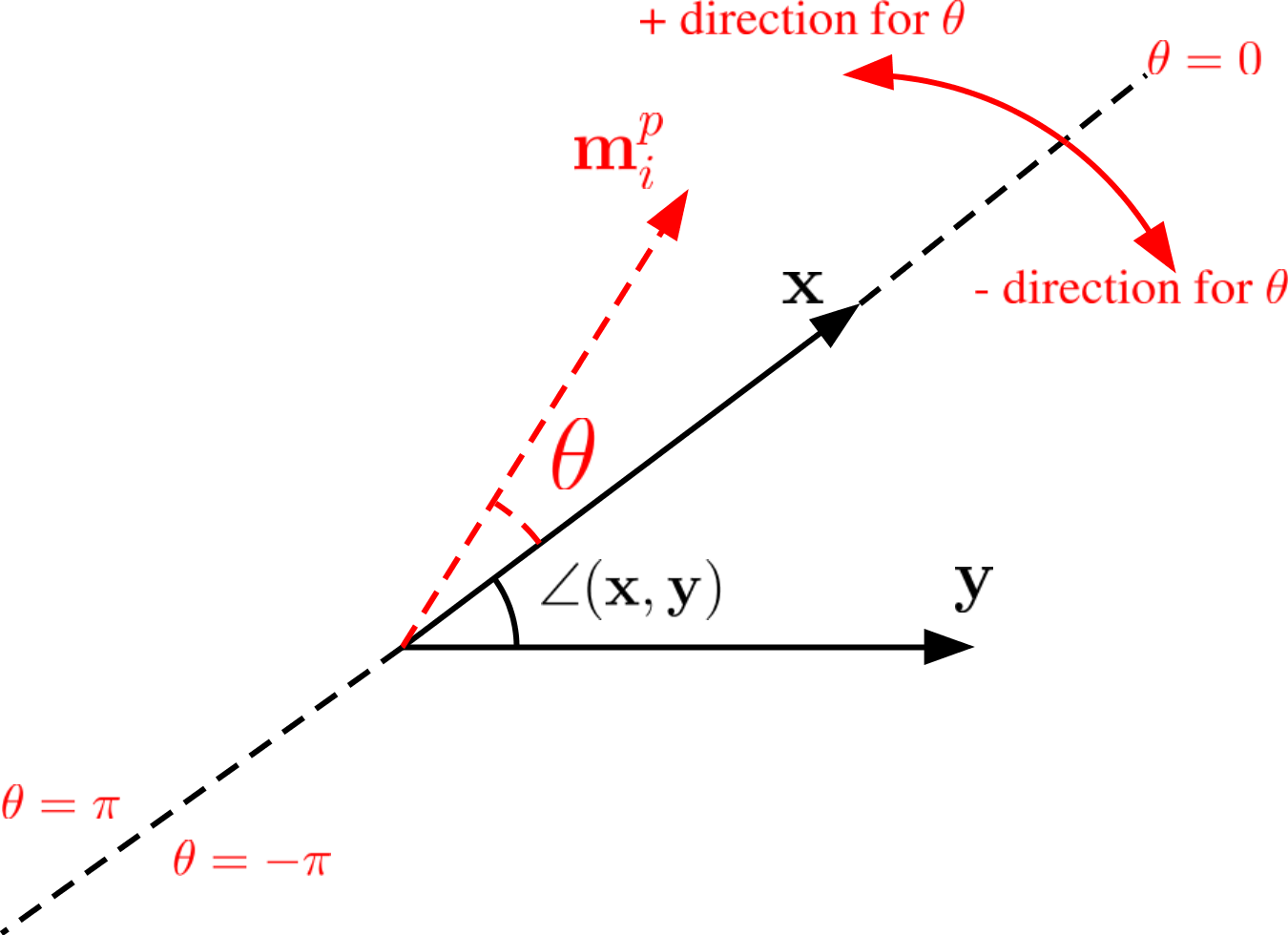}
    \caption{The angle between ${\mathbf m}_i^p$ and ${\mathbf y}$ equals to $\boldsymbol{\theta}+\angle({\mathbf x},{\mathbf y})$,
    $\boldsymbol\theta$ takes values between $-\pi$ and $\pi.$}
    \label{fig1}
\end{figure}

To evaluate the expression ${\mathbb E}[\rho({\mathbf m}_i^T{\mathbf x})\rho({\mathbf m}_i^T{\mathbf y})]$
in \eqref{expe2}, 
we make use of
the fact that the projection ${\mathbf m}_i^p$ of the Gaussian random vector
${\mathbf m}_i$ to the plane spanned by ${\mathbf x}$
and ${\mathbf y}$ is another Gaussian with expected
squared norm $2/m$\textcolor{black}{, as long as ${\mathbf x}\neq{\mathbf y}.$} \textcolor{black}{If the angle between 
${\mathbf m}_i^p$ and ${\mathbf x}$ is 
$\boldsymbol{\theta}\in[-\pi,\pi]$, then the angle between 
${\mathbf m}_i^p$ and ${\mathbf y}$ can be taken as
$\boldsymbol{\theta}+\angle({\mathbf x},{\mathbf y})$
without loss of generality (please see Fig.~\ref{fig1}).}

\textcolor{black}{
In this case we would have}
\begin{align*}
&{\mathbf m}_i^T
{\mathbf x}= ({\mathbf m}_i^p)^T {\mathbf x}= 
\|{\mathbf m}_i^p\|_2
\|{\mathbf x}\|_2\cos\boldsymbol{\theta}\\
&{\mathbf m}_i^T {\mathbf y}= ({\mathbf m}_i^p)^T {\mathbf y} = \|{\mathbf m}_i^p\|_2
\|{\mathbf y}\|_2 \cos(\boldsymbol{\theta}
+\angle({\mathbf x},{\mathbf y})) 
\end{align*}
\textcolor{black}{
The product 
$\rho({\mathbf m}_i^T{\mathbf x})
\rho({\mathbf m}_i^T{\mathbf y})$ is non-zero only when
\begin{equation}
\cos\boldsymbol{\theta}>0, \quad \cos(\boldsymbol{\theta}
+\angle({\mathbf x},{\mathbf y}))>0,\,\, 
\boldsymbol{\theta}\in[-\pi,\pi]\label{cos_ineq}    
\end{equation}
So the inequality
$\cos(\boldsymbol{\theta}+\angle({\mathbf x},{\mathbf y}))>0$
needs to be satisfied under the constraint $\boldsymbol{\theta}\in(-\pi/2,\pi/2).$ Note that
$0\leq\angle({\mathbf x},{\mathbf y})\leq\pi$ by definition,
meaning that \eqref{cos_ineq} implies $\boldsymbol{\theta}+\angle({\mathbf x},{\mathbf y})\in(-\pi/2,3\pi/2).$
But cosine function is non-positive in the interval 
$[\pi/2,3\pi/2].$ Thus the set of conditions \eqref{cos_ineq} 
is satisfied if and only if
%\vspace{-.1in}
\begin{equation*}
-\pi/2<\boldsymbol{\theta}<\pi/2, \quad
-\pi/2<
\boldsymbol{\theta}+
\angle({\mathbf x},{\mathbf y})<\pi/2,
\end{equation*}
which is equivalent to the single inequality given by
%\vspace{-.1in}
$$
-\pi/2<\boldsymbol{\theta}<\pi/2-\angle({\mathbf x},{\mathbf y}).
$$
%\vspace{-.1in}
Thus we can express 
the expectation ${\mathbb E}[\rho({\mathbf m}_i^T{\mathbf x})\rho({\mathbf m}_i^T{\mathbf y})]$ as}
%\vspace{-.05in}
\begin{align}
&\color{black}{{\mathbb E}[\rho({\mathbf m}_i^T{\mathbf x})\rho({\mathbf m}_i^T{\mathbf y})]=\|{\mathbf x}\|_2\|{\mathbf y}\|_2 
{\mathbb E}\|{\mathbf m}_i^p\|^2_2}
\nonumber\\ &\,\color{black}{\cdot
 {\mathbb E}\left[\cos\boldsymbol{\theta}\,
\cos(\boldsymbol{\theta}+\angle({\mathbf x},{\mathbf y}){\mathbbm 1}(-\pi/2<\boldsymbol{\theta}<\pi/2-\angle({\mathbf x},{\mathbf y}))\right]}\nonumber\\
& \color{black}{=
\frac{2\|{\mathbf x}\|_2\|{\mathbf y}\|_2}{m}
\int_{-\pi/2}^{\pi/2-\angle({\mathbf x},{\mathbf y})}
\hspace{-.2in}
\cos\boldsymbol{\theta}\,
\cos(\boldsymbol{\theta}
+\angle({\mathbf x},{\mathbf y}))\,\frac{d\boldsymbol{\theta}}{2\pi}}
\nonumber\\
&=\frac{\|{\mathbf x}\|_2\|{\mathbf y}\|_2}{m\pi}
\nonumber\\&\quad\cdot
\int_{-\pi/2}^{\pi/2-\angle({\mathbf x},{\mathbf y})}
\hspace{-.2in}
\sin\left(\boldsymbol{\theta}+\frac{\pi}{2}\right)
\sin\left(\boldsymbol{\theta}+\frac{\pi}{2}+\angle({\mathbf x},{\mathbf y})\right)\,d\boldsymbol{\theta}
\nonumber\\&
=\frac{\|{\mathbf x}\|_2\|{\mathbf y}\|_2}{m\pi}
\int_{0}^{\pi-\angle({\mathbf x},{\mathbf y})}
\sin\boldsymbol{\theta}\,
\sin(\boldsymbol{\theta}
+\angle({\mathbf x},{\mathbf y}))\,d\boldsymbol{\theta}.
\label{expe5}
\end{align}
\vspace{-.05in}
Inserting the equations \eqref{expe3}--\eqref{expe5} in \eqref{expe2}, we get
\begin{align}
&{\mathbb E}\left(\rho({\mathbf m}_i^T{\mathbf x})-
\rho({\mathbf m}_i^T{\mathbf y})\right)^2=
\frac{1}{m}\bigg(\frac{\|{\mathbf x}\|_2^2}{2}+
\frac{\|{\mathbf y}\|_2^2}{2}
\nonumber\\&\quad
-2
\frac{\|{\mathbf x}\|_2\|{\mathbf y}\|_2}{\pi}
\int_{0}^{\pi-\angle({\mathbf x},{\mathbf y})}
\sin\boldsymbol{\theta}\,
\sin(\boldsymbol{\theta}
+\angle({\mathbf x},{\mathbf y}))\,d\boldsymbol{\theta}
\bigg),\nonumber
\end{align}
\vspace{-.05in}
from which
%\vspace{-.05in}
\begin{align}
&{\mathbb E}\|\rho(\mathbf{Mx})-\rho(\mathbf{My})\|_2^2
=
\frac{\|{\mathbf x}\|_2^2}{2}+
\frac{\|{\mathbf y}\|_2^2}{2}\nonumber\\&\quad
-2
\frac{\|{\mathbf x}\|_2\|{\mathbf y}\|_2}{\pi}
\int_{0}^{\pi-\angle({\mathbf x},{\mathbf y})}
\sin\boldsymbol{\theta}\,
\sin(\boldsymbol{\theta}
+\angle({\mathbf x},{\mathbf y}))\,d\boldsymbol{\theta}
\label{total_exp}
\vspace{-.1in}
\end{align}
follows trivially due to \eqref{expe1}. 
Comparing \eqref{total_exp}
with eq.~(11) in \cite{main}, we observe 
\textcolor{black}{there is a missing multiplicative
factor of $-2$ for the coefficient of the integral term
of eq.~(11) in \cite{main}.}
To evaluate the integral, we write
\begin{align}
&\int_0^{\pi-\angle({\mathbf x},{\mathbf y})}
\sin\boldsymbol{\theta}\,
\sin(\boldsymbol{\theta}
+\angle({\mathbf x},{\mathbf y}))\,d\boldsymbol{\theta}
\nonumber\\
&\hspace{-.15in}=\int_0^{\pi-\angle({\mathbf x},{\mathbf y})}
\left[
\frac{\cos\angle({\mathbf x},{\mathbf y})-
\cos(2\boldsymbol{\theta}+
\angle({\mathbf x},{\mathbf y}))}{2}\right]
\,d\boldsymbol{\theta}\nonumber
\end{align}
\begin{align}
&\quad=
-\frac{\sin(2\boldsymbol{\theta}+
\angle({\mathbf x},{\mathbf y}))}{4}
\bigg|_0^
{\pi-\angle({\mathbf x},{\mathbf y})}\!\!\!\!\!\!+
\frac{\pi-\angle({\mathbf x},{\mathbf
y})}{2}\cos\angle({\mathbf x},{\mathbf y})\nonumber\\
&\quad=\frac{\sin\angle({\mathbf x},{\mathbf y})}{2}
+\frac{\pi-\angle({\mathbf x},{\mathbf
y})}{2}\cos\angle({\mathbf x},{\mathbf y})\nonumber\\
&\quad=\frac{\pi}{2}\cos\angle({\mathbf x},{\mathbf y})
+\frac{1}{2}
\left(\sin\angle({\mathbf x},{\mathbf y})
-\angle({\mathbf x},{\mathbf y})
\cos\angle({\mathbf x},{\mathbf y})
\right).\label{integ}
\end{align}
Note that \textcolor{black}{the right hand side of \eqref{integ} is one half of}
the term in brackets appearing in eq.~(21) of \cite{main},
\textcolor{black}{so they are not equal.}
Combining \eqref{integ} with \eqref{total_exp},
we obtain
\begin{align}
&{\mathbb E}\|\rho(\mathbf{Mx})-\rho(\mathbf{My})\|_2^2
\nonumber\\&\,\,=
\frac{\|{\mathbf x}\|_2^2}{2}+
\frac{\|{\mathbf y}\|_2^2}{2}
-\|{\mathbf x}\|_2\|{\mathbf y}\|_2
\cos\angle({\mathbf x},{\mathbf y})
\nonumber\\&\quad-
\frac{\|{\mathbf x}\|_2\|{\mathbf y}\|_2}{\pi}
\left(\sin\angle({\mathbf x},{\mathbf y})
-\angle({\mathbf x},{\mathbf y})
\cos\angle({\mathbf x},{\mathbf y})
\right)
\nonumber\\&\,\,=\frac{\|{\mathbf x}\|_2^2}{2}+
\frac{\|{\mathbf y}\|_2^2}{2}-{\mathbf x}^T{\mathbf y}-
\|{\mathbf x}\|_2\|{\mathbf y}\|_2
\psi({\mathbf x},{\mathbf y})\label{total_exp2}
\\&\,\,=
\frac{1}{2}\|{\mathbf x}-{\mathbf y}\|_2^2-\|{\mathbf x}\|_2\|{\mathbf y}\|_2
\psi({\mathbf x},{\mathbf y})\label{total_exp3}
\end{align}
where we substitute
$\psi({\mathbf x},{\mathbf y})$ of \eqref{psi}
in \eqref{total_exp2}.

When we compare \eqref{total_exp3} with \eqref{expe}
(or with Eq.~(22) in \cite{main}), we see there is a plus--minus difference associated with the angle--dependent term 
$\|{\mathbf x}\|_2\|{\mathbf y}\|_2
\psi({\mathbf x},{\mathbf y}).$ This difference changes
the conclusion of the derivation fundamentally, i.e.,
it turns out the distance shrinkage that the operator
$\rho({\mathbf M}\cdot)$ induces is greater when the angle between the points is greater. Therefore
the way a DNN with random weights
discriminates angles does not make the classification task
any easier at all, meaning that the classification behavior
illustrated by Fig.~5 in \cite{main} cannot be valid.

Before closing this section, we would like to inform the reader that the usage of Bernstein's inequality and the Gaussian concentration bound in Appendix~A of \cite{main} are correct
to the best of our understanding. Hence following the same lines of reasoning as presented there, it is possible to prove the following.
\begin{theorem}
Let 
$K\subset{\mathbb B}_1^n$ be the manifold of the data
in the input layer.
If $\sqrt{m}{\mathbf M}\in {\mathbb R}^{m\times n}$ is a random matrix with i.i.d. normally distributed entries and if $m$ is sufficiently large (as defined in \cite{main}), then with a high probability
(of the form $1-\exp(-O(\delta^2))$)
\begin{align}
&\Bigg| 
\|\rho(\mathbf{Mx})-
\rho(\mathbf{My})\|_2^2\nonumber\\&\quad-
\left(\frac{1}{2}
\|{\mathbf x}-{\mathbf y}\|_2^2-
\|{\mathbf x}\|_2\|{\mathbf y}\|_2
\psi({\mathbf x},{\mathbf y})\right)\Bigg|\leq \delta.
\label{eq4_corrected}
\end{align}
\label{thm1}
\end{theorem}
\vspace{-.1in}
Theorem \ref{thm1} further corroborates our claim that
DNNs with random weights cannot treat in--class and
out--of--class data in an ideal way. Equation  \eqref{eq4_corrected} implies
inter--class points and intra--class points are treated
in an exactly opposite manner with high probability. 

\section{Angle Distortion and Distance Shrinkage Bound}
\label{cor5}
The angular distance result for a single layer of randomly weighted DNN is provided by Theorem~4 in \cite{main} as 
follows.  
\begin{theorem*}[Theorem 4 in \cite{main}] Under the same
conditions of Theorem~3 in \cite{main} and 
$K\subset {\mathbb B}_1^n\backslash {\mathbb B}_{\beta}^n$,
where $\delta\ll \beta^2<1$, with high probability
\begin{align}
\bigg| \cos
\angle
(\rho({\mathbf M}{\mathbf x}),\rho({\mathbf M}{\mathbf y})) -(\cos\angle({\mathbf x},{\mathbf y})&+
\psi({\mathbf x},{\mathbf y}))\bigg|\nonumber\\&\quad\leq \frac{15\delta}{\beta^2-2\delta}.
\end{align}
\end{theorem*}
This theorem is proved in Appendix~B of \cite{main}. The proof
relies on the inequality
\begin{align}
&\bigg|\frac{1}{2}\|\rho({\mathbf M}{\mathbf x})\|_2^2+
\frac{1}{2}\|\rho({\mathbf M}{\mathbf y})\|_2^2-
\frac{1}{4}
\|{\mathbf x}\|_2^2-\frac{1}{4}
\|{\mathbf y}\|_2^2\nonumber\\
&\quad-\bigg(\rho({\mathbf M}{\mathbf x})^T
\rho({\mathbf M}{\mathbf y})-\frac{\|{\mathbf x}\|_2
\|{\mathbf y}\|_2}{2}
\cos\angle({\mathbf x},{\mathbf y})\nonumber\\
&\quad-\frac{\|{\mathbf x}\|\|{\mathbf y}\|}{2\pi}
(\sin\angle({\mathbf x},{\mathbf y})
-\angle({\mathbf x},{\mathbf y})
\cos\angle({\mathbf x},{\mathbf y}))
\bigg)
\bigg|\leq \frac{\delta}{2}
\label{eq30}
\end{align}
which is given by Eq.~(30) in \cite{main}, where the authors also state that this inequality is equivalent to Eq.~(4)
in \cite{main} (\eqref{eq4} in this work). 

Observing that both 
$\cos\angle({\mathbf x},{\mathbf y}))$ and 
$(\sin\angle({\mathbf x},{\mathbf y})
-\angle({\mathbf x},{\mathbf y})
\cos\angle({\mathbf x},{\mathbf y}))$ terms have the same sign in \eqref{eq30},
it can be seen easily that
\eqref{eq30} (Eq.~(30) in \cite{main}) is not consistent 
with \eqref{eq4} (Eq.~(4) in \cite{main}),
i.e., this part of Appendix~B in \cite{main} seems to be inaccurate. 
In fact, it is relatively straightforward to show that
\eqref{eq30} is equivalent to \eqref{eq4_corrected}.
Hence we conclude \eqref{eq30} is correct 
even if its justification in Appendix~B of
\cite{main} is not.

Consequently Theorem~4 in \cite{main} accurately describes the angular distance behavior for a single DNN layer with random weights.
Note that the experimental results with ImageNet deep network demonstrated by Fig.~4 in \cite{main}
is in full accordance with Theorem~4, as explained in \cite{main}.

Now we turn our attention to the bounds on the
shrinkage of distances given by Corollary~5 in
\cite{main}.
\begin{corollary*}[Corollary~5 in \cite{main}]
Under the same conditions of Theorem~3 in \cite{main}, with high probability(depending on $\delta$ as in
Theorem~3 in \cite{main}) 
\be
\frac{1}{2}\|{\mathbf x}-{\mathbf y}\|_2^2-\delta
\leq 
\|\rho({\mathbf M}{\mathbf x})-
\rho({\mathbf M}{\mathbf y})\|_2^2 \leq
\|{\mathbf x}-{\mathbf y}\|_2^2+\delta
\label{cor5_eq}
\ee
\end{corollary*}
It follows from our Theorem \ref{thm1}  that
\eqref{cor5_eq} needs to be corrected, and its corrected version is given below, along with the
proof. We conclude from Corollary~\ref{cor2}
that the local structure of the data points is preserved with high probability by the transform $\rho({\mathbf M}\cdot).$
\begin{corollary}
Under the same conditions of Theorem~3 in \cite{main}, with high probability(of the form $1-\exp(-O(\delta^2))$)
\be
\frac{1}{4}\|{\mathbf x}-{\mathbf y}\|_2^2-\delta
\leq 
\|\rho({\mathbf M}{\mathbf x})-
\rho({\mathbf M}{\mathbf y})\|_2^2 \leq
\frac{1}{2}\|{\mathbf x}-{\mathbf y}\|_2^2+\delta
\label{cor5_eq_corrected}
\ee
\label{cor2}
\end{corollary}
\vspace{-.2in}
\begin{proof}
Equation \eqref{eq4_corrected} implies it is sufficient to prove
\begin{align}
\frac{1}{4}\|{\mathbf x}-{\mathbf y}\|_2^2\leq
\frac{1}{2}\|{\mathbf x}-{\mathbf y}\|_2^2
-\|{\mathbf x}\|_2 \|{\mathbf y}\|_2 \psi
({\mathbf x},{\mathbf y}) \leq
\frac{1}{2}\|{\mathbf x}-{\mathbf y}\|_2^2
\label{pf1}
\end{align}
\vspace{-.1in}
Since $\psi({\mathbf x},{\mathbf y})$ is non-negative,
it is trivial to see 
\be
\frac{1}{2}\|{\mathbf x}-{\mathbf y}\|_2^2
-\|{\mathbf x}\|_2 \|{\mathbf y}\|_2 \psi
({\mathbf x},{\mathbf y}) \leq
\frac{1}{2}\|{\mathbf x}-{\mathbf y}\|_2^2.
\ee
To prove the lower bound in \eqref{pf1}, we write
\begin{align}
&\frac{1}{2}\|{\mathbf x}-{\mathbf y}\|_2^2
-\|{\mathbf x}\|_2 \|{\mathbf y}\|_2 \psi
({\mathbf x},{\mathbf y})
\nonumber\\&\,\,
=\frac{1}{2}\|{\mathbf x}-{\mathbf y}\|_2^2
\nonumber\\&\quad
-
\frac{\|{\mathbf x}\|_2 \|{\mathbf y}\|_2}{\pi}
(\sin\angle({\mathbf x},{\mathbf y})-
\angle({\mathbf x},{\mathbf y})\cos\angle({\mathbf x},{\mathbf y}))
\nonumber\\&\,\,
\geq \frac{1}{2}\|{\mathbf x}-{\mathbf y}\|_2^2
+\left(
\frac{\cos\angle({\mathbf x},{\mathbf y})-1}{2}\right)
\|{\mathbf x}\|_2 \|{\mathbf y}\|_2\label{pf2}
\\&\,\,=\frac{{\mathbf x}^T{\mathbf x}}{2}
+\frac{{\mathbf y}^T{\mathbf y}}{2}
-\frac{{\mathbf x}^T{\mathbf y}}{2}
-\frac{\|{\mathbf x}\|_2 \|{\mathbf y}\|_2}{2}
\nonumber\\&\,\,
=\frac{1}{4}\|{\mathbf x}-{\mathbf y}\|_2^2+
\frac{1}{4}(\|{\mathbf x}\|_2-\|{\mathbf y}\|_2)^2
\nonumber\\&\,\,
\geq \frac{1}{4}\|{\mathbf x}-{\mathbf y}\|_2^2
\end{align}
where we have \eqref{pf2} because of
$\psi({\mathbf x},{\mathbf y})$ being a convex 
function of $\cos\angle({\mathbf x},{\mathbf y}).$
(Note that $\psi({\mathbf x},{\mathbf y})=\frac{1}{\pi}
(\sqrt{1-\cos^2\angle({\mathbf x},{\mathbf y})}-
\cos^{-1}(\cos\angle({\mathbf x},{\mathbf y}))\cos\angle({\mathbf x},{\mathbf y}))$ and
the second derivative of $f(t)=\frac{1}{\pi}(\sqrt{1-t^2}-t\cos^{-1}(t))$
is $f''(t)=\frac{1}{\pi\sqrt{1-t^2}}>0.$)
\end{proof}
\vspace{.1in}
\section{Discussions and Conclusions}
\label{concl}
For a single DNN layer with random Gaussian weights,
we have seen in Theorem~\ref{thm1} that the separation of
classes with distinguishable angles becomes more difficult,
and proved in Corollary~\ref{cor2} that the metric structure of the data manifold is not altered. It is possible to extend those results 
along with the angular distortion result given by
Theorem~4 in \cite{main} to the entire network by resorting to the covering number arguments of Section~IV (particularly Theorem~6) in \cite{main}.

In fact, our findings in regard to the angular separation inability of randomly weighted DNNs are supported by the experimental results in Section~VI of \cite{main}. We see from Figs.~6(a)--(b) and
Figs.~7(a)--(b) in \cite{main} that  closest 
inter--class Euclidean distances decrease(the blue curve in Fig.6(a) is biased below 1) and farthest
intra--class Euclidean distances increase for 
CIFAR--10 dataset(the blue curve in Fig.6(b) is biased above 1). This behavior is in strict contrast with the performance of the trained networks and with what a ``universal classifier'' that Theorem~3 in \cite{main} predicts is supposed to do. Theorem~\ref{thm1} we present here explains this discrepancy very well. Similar comments
apply to Figs.~10(a)--(b) and Figs.~11(a)--(b) in
\cite{main},
where we observe for ImageNet dataset that closest inter-class Euclidean distances do shrink but farthest intra-class Euclidean distances do not.

The experiments considered by \cite{main}
is consistent with the existence of distance lower and upper bounds we state in Corollary~\ref{cor2}
as well. 
For a randomly weighted DNN,
Figs.~8(a)--(b), Figs.~9(a)--(b) and Figs.~12(a)--(b),
Figs.~13(a)--(b) in \cite{main} demonstrate that
there is no difference between inter-class and intra-class points in terms of the distance ratios
for CIFAR--10 dataset and ImageNet dataset, respectively. Those results confirm Corollary~\ref{cor2} since the distance bounds in this corollary are valid for both intra-class and inter-class points.

We know that a complete and profound theoretical
explanation for the practical performance of DNNs is still unavailable. Analysis and properties of DNNs involving randomness or random weights are considered by a significant number of papers including %\cite{Rahimi09,Jarrett09,Saxe11,Arora14,
%Choromanska15,Penn17_1,Penn17_2}
\cite{Rahimi09}--\cite{Penn17_2}.
We hope this work will shed some light on 
random matrix theory based approaches and initiate 
some rethinking, perhaps similarly to the way
\cite{Zhang16} contributed to the literature on the
generalization subject.

\ifCLASSOPTIONcaptionsoff
  \newpage
\fi

\remove{
\begin{IEEEbiography}[{\includegraphics[width=1in,height=1.25in,clip,keepaspectratio]{A1.jpg}}]{Talha Cihad Gulcu}
(S'13) received the B.S. degree from Middle East Technical University, Ankara, in 2009, the M.S. degree from Bilkent University, Ankara, in 2011, and the Ph.D. degree from University of Maryland, College Park, MD, in 2015, all in electrical engineering. His research interests are information theory and signal processing for communications.
\end{IEEEbiography}
}

% that's all folks
\end{document}